\documentclass[10pt,twocolumn,letterpaper]{article}

\usepackage{iccv}
\usepackage{rotating}
\usepackage{times}
\usepackage{amsbsy}
\usepackage{amstext}
\usepackage{graphicx}
\usepackage[pagebackref=true,breaklinks=true,letterpaper=true,colorlinks,bookmarks=false]{hyperref}

\iccvfinalcopy 

\def\iccvPaperID{****} 

\providecommand{\tabularnewline}{\\}

\usepackage{url}
\usepackage{booktabs}
\usepackage{pifont}
\usepackage{amssymb}

\ificcvfinal\fi
\begin{document}

\title{An End-to-End Trainable Neural Network for Image-based Sequence Recognition and Its Application to Scene Text Recognition}

\author{Baoguang Shi, Xiang Bai and Cong Yao\\
    School of Electronic Information and Communications\\
    Huazhong University of Science and Technology, Wuhan, China\\
    {\tt\small \{shibaoguang,xbai\}@hust.edu.cn, yaocong2010@gmail.com}
}

\maketitle

\begin{abstract}
Image-based sequence recognition has been a long-standing research
topic in computer vision. In this paper, we investigate the problem
of scene text recognition, which is among the most important and challenging
tasks in image-based sequence recognition. A novel neural network
architecture, which integrates feature extraction, sequence modeling
and transcription into a unified framework, is proposed. Compared
with previous systems for scene text recognition, the proposed architecture
possesses four distinctive properties: (1) It is end-to-end trainable,
in contrast to most of the existing algorithms whose components are
separately trained and tuned. (2) It naturally handles sequences in
arbitrary lengths, involving no character segmentation or horizontal
scale normalization. (3) It is not confined to any predefined lexicon
and achieves remarkable performances in both lexicon-free and lexicon-based
scene text recognition tasks. (4) It generates an effective yet much
smaller model, which is more practical for real-world application
scenarios. The experiments on standard benchmarks, including the IIIT-5K,
Street View Text and ICDAR datasets, demonstrate the superiority of
the proposed algorithm over the prior arts. Moreover, the proposed
algorithm performs well in the task of image-based music score recognition,
which evidently verifies the generality of it.
\end{abstract}

\section{Introduction}

Recently, the community has seen a strong revival
of neural networks, which is mainly stimulated by the great success
of deep neural network models, specifically Deep Convolutional Neural
Networks (DCNN), in various vision tasks. However, majority of the
recent works related to deep neural networks have devoted to detection
or classification of object categories \cite{GirshickDDM14,KrizhevskySH12}.
In this paper, we are concerned with a classic problem in computer
vision: image-based sequence recognition. In real world, a stable
of visual objects, such as scene text, handwriting and musical score,
tend to occur in the form of sequence, not in isolation. Unlike general
object recognition, recognizing such sequence-like objects often requires
the system to predict a series of object labels, instead of a single
label. Therefore, recognition of such objects can be naturally cast
as a sequence recognition problem. Another unique property of sequence-like
objects is that their lengths may vary drastically. For instance,
English words can either consist of 2 characters such as ``OK''
or 15 characters such as ``congratulations''. Consequently, the
most popular deep models like DCNN \cite{KrizhevskySH12,LecunLYP1998}
cannot be directly applied to sequence prediction, since DCNN models
often operate on inputs and outputs with fixed dimensions, and thus
are incapable of producing a variable-length label sequence.

Some attempts have been made to address this problem for a specific
sequence-like object (\emph{e.g.} scene text). For example, the algorithms
in \cite{WangWCN12,BissaccoCNN13} firstly detect individual characters
and then recognize these detected characters with DCNN models, which
are trained using labeled character images. Such methods often require
training a strong character detector for accurately detecting and
cropping each character out from the original word image. Some other
approaches (such as \cite{JaderbergSVZ14a}) treat scene text recognition
as an image classification problem, and assign a class label to each
English word (90K words in total). It turns out a large trained model
with a huge number of classes, which is difficult to be generalized
to other types of sequence-like objects, such as Chinese texts, musical
scores, \emph{etc.}, because the numbers of basic combinations of
such kind of sequences can be greater than 1 million. In summary,
current systems based on DCNN can not be directly used for image-based
sequence recognition.

Recurrent neural networks (RNN) models, another important branch of
the deep neural networks family, were mainly designed for handling
sequences. One of the advantages of RNN is that it does not need the
position of each element in a sequence object image in both training
and testing. However, a preprocessing step that converts an input
object image into a sequence of image features, is usually essential.
For example, Graves \emph{et al.} \cite{GravesLFBBS09} extract a
set of geometrical or image features from handwritten texts, while
Su and Lu \cite{SuL14} convert word images into sequential HOG features.
The preprocessing step is independent of the subsequent components
in the pipeline, thus the existing systems based on RNN can not be
trained and optimized in an end-to-end fashion.

Several conventional scene text recognition methods that are not based
on neural networks also brought insightful ideas and novel representations
into this field. For example, Almaz\`an \emph{et al.~}\cite{AlmazanGFV14}
and Rodriguez-Serrano \emph{et al.}~\cite{Rodriguez-Serrano15} proposed
to embed word images and text strings in a common vectorial subspace,
and word recognition is converted into a retrieval problem. Yao \emph{et
al.}~\cite{YaoBSL14} and Gordo \emph{et al.}~\cite{Gordo14} used
mid-level features for scene text recognition. Though achieved promising
performance on standard benchmarks, these methods are generally outperformed
by previous algorithms based on neural networks~\cite{BissaccoCNN13,JaderbergSVZ14a},
as well as the approach proposed in this paper.

The main contribution of this paper is a novel neural network model,
whose network architecture is specifically designed for recognizing
sequence-like objects in images. The proposed neural network model
is named as Convolutional Recurrent Neural Network (CRNN), since it
is a combination of DCNN and RNN. For sequence-like objects, CRNN
possesses several distinctive advantages over conventional neural
network models: 1) It can be directly learned from sequence labels
(for instance, words), requiring no detailed annotations (for instance,
characters); 2) It has the same property of DCNN on learning informative
representations directly from image data, requiring neither hand-craft
features nor preprocessing steps, including binarization/segmentation,
component localization, \emph{etc.}; 3) It has the same property of
RNN, being able to produce a sequence of labels; 4) It is unconstrained
to the lengths of sequence-like objects, requiring only height normalization
in both training and testing phases; 5) It achieves better or highly
competitive performance on scene texts (word recognition) than the
prior arts \cite{JaderbergVZ14,BissaccoCNN13}; 6) It contains much
less parameters than a standard DCNN model, consuming less storage
space.

\section{The Proposed Network Architecture}

\begin{figure}[h]
\begin{centering}
\includegraphics[width=0.9\linewidth]{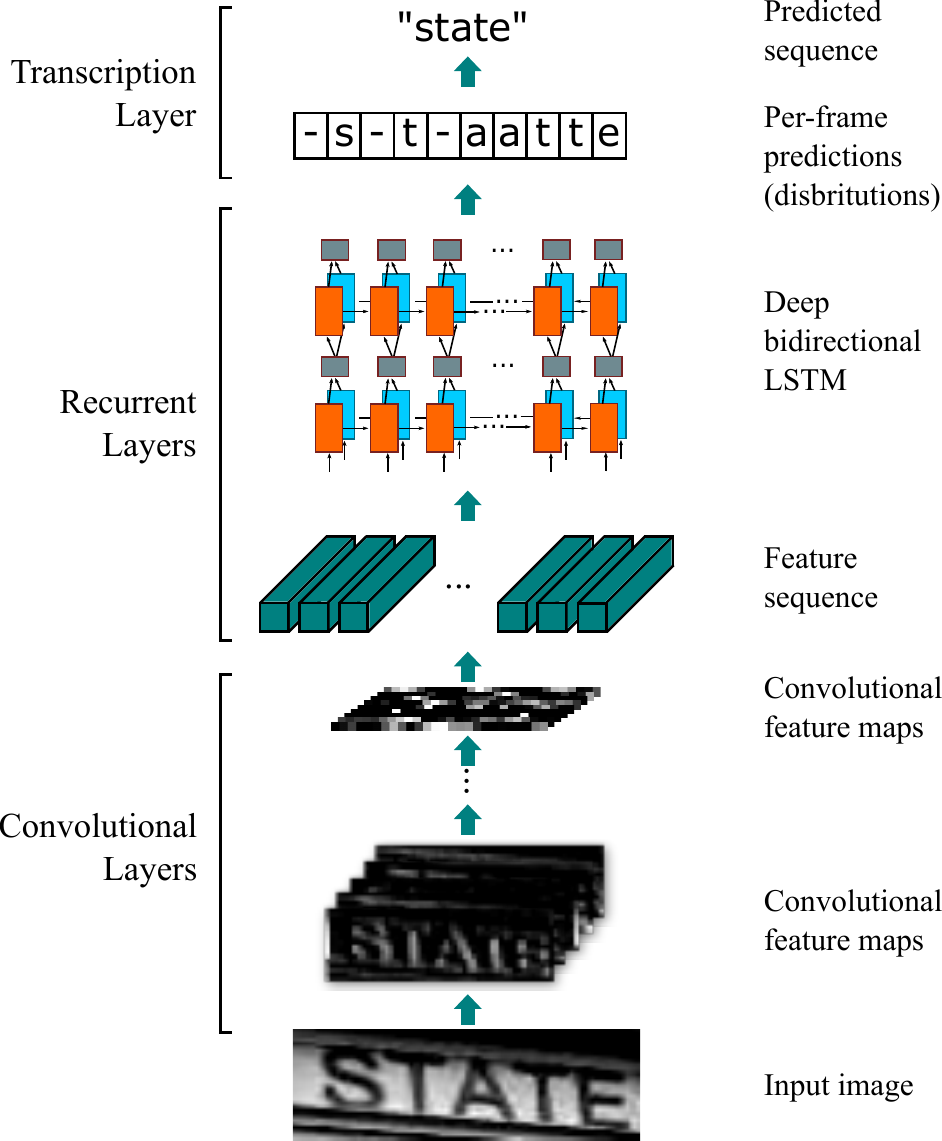}
\par\end{centering}

\caption{The network architecture. The architecture consists of three parts:
1) convolutional layers, which extract a feature sequence from the
input image; 2) recurrent layers, which predict a label distribution
for each frame; 3) transcription layer, which translates the per-frame
predictions into the final label sequence.}
\label{fig:netarch}
\end{figure}

The network architecture of CRNN, as shown in Fig.~\ref{fig:netarch},
consists of three components, including the convolutional layers,
the recurrent layers, and a transcription layer, from bottom to top.

At the bottom of CRNN, the convolutional layers automatically extract
a feature sequence from each input image. On top of the convolutional
network, a recurrent network is built for making prediction for each
frame of the feature sequence, outputted by the convolutional layers.
The transcription layer at the top of CRNN is adopted to translate
the per-frame predictions by the recurrent layers into a label sequence.
Though CRNN is composed of different kinds of network architectures
(eg. CNN and RNN), it can be jointly trained with one loss function.

\subsection{Feature Sequence Extraction}

In CRNN model, the component of convolutional layers is constructed
by taking the convolutional and max-pooling layers from a standard
CNN model (fully-connected layers are removed). Such component is
used to extract a sequential feature representation from an input
image. Before being fed into the network, all the images need to be
scaled to the same height. Then a sequence of feature vectors is extracted
from the feature maps produced by the component of convolutional layers,
which is the input for the recurrent layers. Specifically, each feature
vector of a feature sequence is generated from left to right on the
feature maps by column. This means the $i$-th feature vector is the
concatenation of the $i$-th columns of all the maps. The width of
each column in our settings is fixed to single pixel.

\begin{figure}[h]
\begin{centering}
\includegraphics[width=0.45\linewidth]{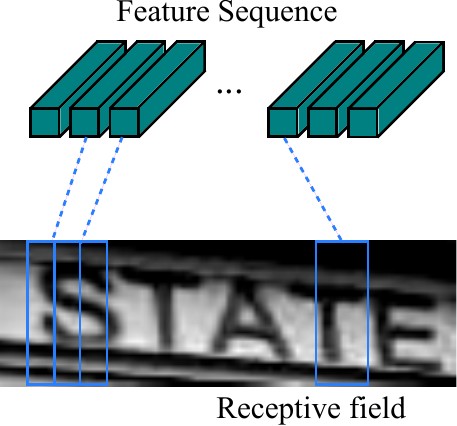}
\par\end{centering}

\caption{The receptive field. Each vector in the extracted feature sequence
is associated with a receptive field on the input image, and can be
considered as the feature vector of that field.}

\label{fig:recfield}
\end{figure}

As the layers of convolution, max-pooling, and element-wise activation
function operate on local regions, they are translation invariant.
Therefore, each column of the feature maps corresponds to a rectangle
region of the original image (termed the \emph{receptive field}),
and such rectangle regions are in the same order to their corresponding
columns on the feature maps from left to right. As illustrated in
Fig.~\ref{fig:recfield}, each vector in the feature sequence is
associated with a receptive field, and can be considered as the image
descriptor for that region.

Being robust, rich and trainable, deep convolutional features have
been widely adopted for different kinds of visual recognition tasks~\cite{KrizhevskySH12,GirshickDDM14}.
Some previous approaches have employed CNN to learn a robust representation
for sequence-like objects such as scene text \cite{JaderbergSVZ14a}.
However, these approaches usually extract holistic representation
of the whole image by CNN, then the local deep features are collected
for recognizing each component of a sequence-like object. Since CNN
requires the input images to be scaled to a fixed size in order to
satisfy with its fixed input dimension, it is not appropriate for
sequence-like objects due to their large length variation. In CRNN,
we convey deep features into sequential representations in order to
be invariant to the length variation of sequence-like objects.

\subsection{Sequence Labeling}

\begin{figure}
\begin{centering}
\includegraphics[width=1\linewidth]{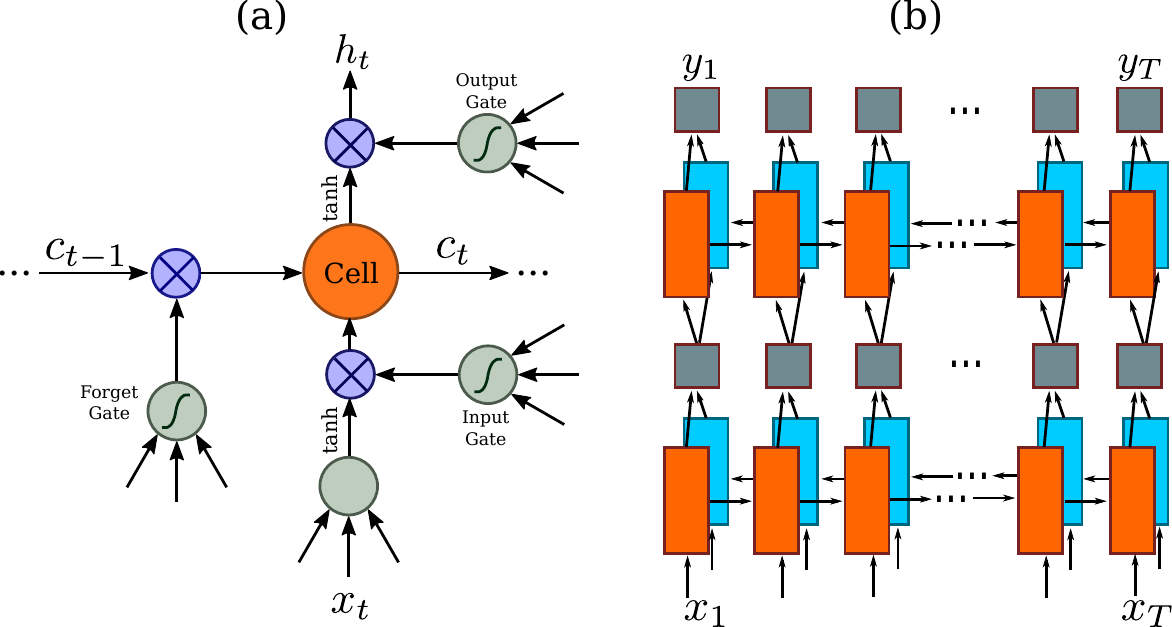}
\par\end{centering}

\caption{(a) The structure of a basic LSTM unit. An LSTM consists of a cell
module and three gates, namely the input gate, the output gate and
the forget gate. (b) The structure of deep bidirectional LSTM we use
in our paper. Combining a forward (left to right) and a backward (right
to left) LSTMs results in a bidirectional LSTM. Stacking multiple
bidirectional LSTM results in a deep bidirectional LSTM.}

\label{fig:lstm}
\end{figure}

A deep bidirectional Recurrent Neural Network is built on the top
of the convolutional layers, as the recurrent layers. The recurrent
layers predict a label distribution $y_{t}$ for each frame $x_{t}$
in the feature sequence $\mathbf{x}=x_{1},\dots,x_{T}$. The advantages
of the recurrent layers are three-fold. Firstly, RNN has a strong
capability of capturing contextual information within a sequence.
Using contextual cues for image-based sequence recognition is more
stable and helpful than treating each symbol independently. Taking
scene text recognition as an example, wide characters may require
several successive frames to fully describe (refer to Fig.~\ref{fig:recfield}).
Besides, some ambiguous characters are easier to distinguish when
observing their contexts, \emph{e.g.} it is easier to recognize ``il''
by contrasting the character heights than by recognizing each of them
separately. Secondly, RNN can back-propagates error differentials
to its input, \emph{i.e.} the convolutional layer, allowing us to
jointly train the recurrent layers and the convolutional layers in
a unified network. Thirdly, RNN is able to operate on sequences of
arbitrary lengths, traversing from starts to ends.

A traditional RNN unit has a self-connected hidden layer between its
input and output layers. Each time it receives a frame $x_{t}$ in
the sequence, it updates its internal state $h_{t}$ with a non-linear
function that takes both current input $x_{t}$ and past state $h_{t-1}$
as its inputs: $h_{t}=g(x_{t},h_{t-1})$. Then the prediction $y_{t}$
is made based on $h_{t}$. In this way, past contexts $\{x_{t'}\}_{t'<t}$
are captured and utilized for prediction. Traditional RNN unit, however,
suffers from the vanishing gradient problem~\cite{BengioSF94}, which
limits the range of context it can store, and adds burden to the training
process. Long-Short Term Memory~\cite{HochreiterS97,GersSS02} (LSTM)
is a type of RNN unit that is specially designed to address this problem.
An LSTM (illustrated in Fig.~\ref{fig:lstm}) consists of a memory
cell and three multiplicative gates, namely the input, output and
forget gates. Conceptually, the memory cell stores the past contexts,
and the input and output gates allow the cell to store contexts for
a long period of time. Meanwhile, the memory in the cell can be cleared
by the forget gate. The special design of LSTM allows it to capture
long-range dependencies, which often occur in image-based sequences.

LSTM is directional, it only uses past contexts. However, in image-based
sequences, contexts from both directions are useful and complementary
to each other. Therefore, we follow \cite{GravesMH13} and combine
two LSTMs, one forward and one backward, into a bidirectional LSTM.
Furthermore, multiple bidirectional LSTMs can be stacked, resulting
in a deep bidirectional LSTM as illustrated in Fig.~\ref{fig:lstm}.b.
The deep structure allows higher level of abstractions than a shallow
one, and has achieved significant performance improvements in the
task of speech recognition \cite{GravesMH13}.

In recurrent layers, error differentials are propagated in the opposite
directions of the arrows shown in Fig.~\ref{fig:lstm}.b, \emph{i.e.}
Back-Propagation Through Time (BPTT). At the bottom of the recurrent
layers, the sequence of propagated differentials are concatenated
into maps, inverting the operation of converting feature maps into
feature sequences, and fed back to the convolutional layers. In practice,
we create a custom network layer, called ``Map-to-Sequence'', as
the bridge between convolutional layers and recurrent layers.

\subsection{Transcription}

Transcription is the process of converting the per-frame predictions
made by RNN into a label sequence. Mathematically, transcription is
to find the label sequence with the highest probability conditioned
on the per-frame predictions. In practice, there exists two modes
of transcription, namely the lexicon-free and lexicon-based transcriptions.
A lexicon is a set of label sequences that prediction is constraint
to, \emph{e.g.} a spell checking dictionary. In lexicon-free mode,
predictions are made without any lexicon. In lexicon-based mode, predictions
are made by choosing the label sequence that has the highest probability.

\subsubsection{Probability of label sequence\label{sec:stringprob}}

We adopt the conditional probability defined in the Connectionist
Temporal Classification (CTC) layer proposed by Graves \emph{et al.}
\cite{GravesFGS06}. The probability is defined for label sequence
$\mathbf{l}$ conditioned on the per-frame predictions $\mathbf{y}=y_{1},\dots,y_{T}$,
and it ignores the position where each label in $\mathbf{l}$ is located.
Consequently, when we use the negative log-likelihood of this probability
as the objective to train the network, we only need images and their
corresponding label sequences, avoiding the labor of labeling positions
of individual characters.

The formulation of the conditional probability is briefly described
as follows: The input is a sequence $\mathbf{y}=y_{1},\dots,y_{T}$
where $T$ is the sequence length. Here, each $y_{t}\in\Re^{|{\cal L}'|}$
is a probability distribution over the set ${\cal L}'={\cal L}\cup\textrm{�}$,
where ${\cal L}$ contains all labels in the task (\emph{e.g.} all
English characters), as well as a 'blank' label denoted by $\textrm{�}$.
A sequence-to-sequence mapping function ${\cal B}$ is defined on
sequence $\boldsymbol{\pi}\in{\cal L}'^{T}$, where $T$ is the length.
${\cal B}$ maps $\boldsymbol{\pi}$ onto $\mathbf{l}$ by firstly
removing the repeated labels, then removing the 'blank's. For example,
${\cal B}$ maps ``\texttt{-{}-hh-e-l-ll-oo-{}-}'' ('\texttt{-}'
represents 'blank') onto ``\texttt{hello}''. Then, the conditional
probability is defined as the sum of probabilities of all $\boldsymbol{\pi}$
that are mapped by ${\cal B}$ onto $\mathbf{l}$:

\begin{equation}
p(\mathbf{l}|\mathbf{y})=\sum_{\boldsymbol{\pi}:{\cal B}(\boldsymbol{\pi})=\mathbf{l}}p(\boldsymbol{\pi}|\mathbf{y}),\label{eq:stringprob}
\end{equation}
where the probability of $\boldsymbol{\pi}$ is defined as $p(\boldsymbol{\pi}|\mathbf{y})=\prod_{t=1}^{T}y_{\pi_{t}}^{t}$,
$y_{\pi_{t}}^{t}$ is the probability of having label $\pi_{t}$ at
time stamp $t$. Directly computing Eq.~\ref{eq:stringprob} would
be computationally infeasible due to the exponentially large number
of summation items. However, Eq.~\ref{eq:stringprob} can be efficiently
computed using the forward-backward algorithm described in~\cite{GravesFGS06}.

\subsubsection{Lexicon-free transcription\label{sec:lexfree}}

In this mode, the sequence $\mathbf{l}^{*}$ that has the highest
probability as defined in Eq.~\ref{eq:stringprob} is taken as the
prediction. Since there exists no tractable algorithm to precisely
find the solution, we use the strategy adopted in \cite{GravesFGS06}.
The sequence $\mathbf{l}^{*}$ is approximately found by $\mathbf{l}^{*}\approx{\cal B}(\arg\max_{\boldsymbol{\pi}}p(\boldsymbol{\pi}|\mathbf{y}))$,
\emph{i.e.} taking the most probable label $\pi_{t}$ at each time
stamp $t$, and map the resulted sequence onto $\mathbf{l}^{*}$.

\subsubsection{Lexicon-based transcription\label{sec:lexbased}}

In lexicon-based mode, each test sample is associated with a lexicon
${\cal D}$. Basically, the label sequence is recognized by choosing
the sequence in the lexicon that has highest conditional probability
defined in Eq.~\ref{eq:stringprob}, \emph{i.e.} $\mathbf{l}^{*}=\arg\max_{\mathbf{l}\in{\cal D}}p(\mathbf{l}|\mathbf{y})$.
However, for large lexicons\emph{, e.g.} the 50k-words Hunspell spell-checking
dictionary~\cite{Hunspell}, it would be very time-consuming to perform
an exhaustive search over the lexicon, \emph{i.e.} to compute Equation~\ref{eq:stringprob}
for all sequences in the lexicon and choose the one with the highest
probability. To solve this problem, we observe that the label sequences
predicted via lexicon-free transcription, described in \ref{sec:lexfree},
are often close to the ground-truth under the edit distance metric.
This indicates that we can limit our search to the nearest-neighbor
candidates ${\cal N}_{\delta}(\mathbf{l}')$, where $\delta$ is the
maximal edit distance and $\mathbf{l}'$ is the sequence transcribed
from $\mathbf{y}$ in lexicon-free mode:

\begin{equation}
\mathbf{l}^{*}=\arg\max_{\mathbf{l}\in{\cal N}_{\delta}(\mathbf{l}')}p(\mathbf{l}|\mathbf{y}).\label{eq:largelex}
\end{equation}

The candidates ${\cal N}_{\delta}(\mathbf{l}')$ can be found efficiently
with the BK-tree data structure~\cite{BurkhardK73}, which is a metric
tree specifically adapted to discrete metric spaces. The search time
complexity of BK-tree is $O(\log|{\cal D}|)$, where $|{\cal D}|$
is the lexicon size. Therefore this scheme readily extends to very
large lexicons. In our approach, a BK-tree is constructed offline
for a lexicon. Then we perform fast online search with the tree, by
finding sequences that have less or equal to $\delta$ edit distance
to the query sequence.

\subsection{Network Training\label{sec:nettrain}}

Denote the training dataset by ${\cal X}=\{I_{i},\mathbf{l}_{i}\}_{i}$,
where $I_{i}$ is the training image and $\mathbf{l}_{i}$ is the
ground truth label sequence. The objective is to minimize the negative
log-likelihood of conditional probability of ground truth:

\begin{equation}
{\cal O}=-\sum_{I_{i},\mathbf{l}_{i}\in{\cal X}}\log p(\mathbf{l}_{i}|\mathbf{y}_{i}),\label{eq:objective}
\end{equation}
where $\mathbf{y}_{i}$ is the sequence produced by the recurrent
and convolutional layers from $I_{i}$. This objective function calculates
a cost value directly from an image and its ground truth label sequence.
Therefore, the network can be end-to-end trained on pairs of images
and sequences, eliminating the procedure of manually labeling all
individual components in training images.

The network is trained with stochastic gradient descent (SGD). Gradients
are calculated by the back-propagation algorithm. In particular, in
the transcription layer, error differentials are back-propagated with
the forward-backward algorithm, as described in~\cite{GravesFGS06}.
In the recurrent layers, the Back-Propagation Through Time (BPTT)
is applied to calculate the error differentials.

For optimization, we use the ADADELTA~\cite{Matthew12ADADELTA} to
automatically calculate per-dimension learning rates. Compared with
the conventional momentum \cite{LearningRepresentations} method,
ADADELTA requires no manual setting of a learning rate. More importantly,
we find that optimization using ADADELTA converges faster than the
momentum method.

\section{Experiments}

To evaluate the effectiveness of the proposed CRNN model, we conducted
experiments on standard benchmarks for scene text recognition and
musical score recognition, which are both challenging vision tasks.
The datasets and setting for training and testing are given in Sec.~\ref{sec:datasets},
the detailed settings of CRNN for scene text images is provided in
Sec.~\ref{sec:impldetails}, and the results with the comprehensive
comparisons are reported in Sec.~\ref{sec:evaluation}. To further
demonstrate the generality of CRNN, we verify the proposed algorithm
on a music score recognition task in Sec.~\ref{sec:musicalscore}.

\subsection{Datasets\label{sec:datasets}}

For all the experiments for scene text recognition, we use the synthetic
dataset (Synth) released by Jaderberg \emph{et al.} \cite{JaderbergSVZ14}
as the training data. The dataset contains 8 millions training images
and their corresponding ground truth words. Such images are generated
by a synthetic text engine and are highly realistic. Our network is
trained on the synthetic data once, and tested on all other real-world
test datasets without any fine-tuning on their training data. Even
though the CRNN model is purely trained with synthetic text data,
it works well on real images from standard text recognition benchmarks.

Four popular benchmarks for scene text recognition are used for performance
evaluation, namely ICDAR~2003 (IC03), ICDAR~2013 (IC13), IIIT~5k-word
(IIIT5k), and Street View Text (SVT).

\textbf{IC03} \cite{LucasPSTWYANOYMZOWJTWL05} test dataset contains
251 scene images with labeled text bounding boxes. Following Wang
\emph{et al.} \cite{WangBB11}, we ignore images that either contain
non-alphanumeric characters or have less than three characters, and
get a test set with 860 cropped text images. Each test image is associated
with a 50-words lexicon which is defined by Wang \emph{et al.} \cite{WangBB11}.
A full lexicon is built by combining all the per-image lexicons. In
addition, we use a 50k words lexicon consisting of the words in the
Hunspell spell-checking dictionary \cite{Hunspell}.

\textbf{IC13} \cite{KaratzasSUIBMMMAH13} test dataset inherits most
of its data from IC03. It contains 1,015 ground truths cropped word
images.

\textbf{IIIT5k \cite{MishraAJ12}} contains 3,000 cropped word test
images collected from the Internet. Each image has been associated
to a 50-words lexicon and a 1k-words lexicon.

\textbf{SVT \cite{WangBB11}} test dataset consists of 249 street
view images collected from Google Street View. From them 647 word
images are cropped. Each word image has a 50 words lexicon defined
by Wang \emph{et al.} \cite{WangBB11}.

\begin{table}
\caption{Network configuration summary. The first row is the top layer.
`k', `s' and `p' stand for kernel size, stride and padding size respectively}

\footnotesize

\begin{centering}
\begin{tabular}{|c|c|}
\hline 
\textbf{Type} & \textbf{Configurations}\tabularnewline
\hline 
\hline 
Transcription & -\tabularnewline
\hline 
Bidirectional-LSTM & \#hidden units:256\tabularnewline
\hline 
Bidirectional-LSTM & \#hidden units:256\tabularnewline
\hline 
Map-to-Sequence & -\tabularnewline
\hline 
Convolution & \#maps:512, k:$2\times2$, s:1, p:0\tabularnewline
\hline 
MaxPooling & Window:$1\times2$, s:2\tabularnewline
\hline 
BatchNormalization & -\tabularnewline
\hline 
Convolution & \#maps:512, k:$3\times3$, s:1, p:1\tabularnewline
\hline 
BatchNormalization & -\tabularnewline
\hline 
Convolution & \#maps:512, k:$3\times3$, s:1, p:1\tabularnewline
\hline 
MaxPooling & Window:$1\times2$, s:2\tabularnewline
\hline 
Convolution & \#maps:256, k:$3\times3$, s:1, p:1\tabularnewline
\hline 
Convolution & \#maps:256, k:$3\times3$, s:1, p:1\tabularnewline
\hline 
MaxPooling & Window:$2\times2$, s:2\tabularnewline
\hline 
Convolution & \#maps:128, k:$3\times3$, s:1, p:1\tabularnewline
\hline 
MaxPooling & Window:$2\times2$, s:2\tabularnewline
\hline 
Convolution & \#maps:64, k:$3\times3$, s:1, p:1\tabularnewline
\hline 
Input & $W\times32$ gray-scale image\tabularnewline
\hline 
\end{tabular}
\par\end{centering}

\label{tbl:netconfig}
\end{table}

\subsection{Implementation Details\label{sec:impldetails}}

The network configuration we use in our experiments is summarized
in Table~\ref{tbl:netconfig}. The architecture of the convolutional
layers is based on the VGG-VeryDeep architectures~\cite{SimonyanZ14a}.
A tweak is made in order to make it suitable for recognizing English
texts. In the 3rd and the 4th max-pooling layers, we adopt $1\times2$
sized rectangular pooling windows instead of the conventional squared
ones. This tweak yields feature maps with larger width, hence longer
feature sequence. For example, an image containing 10 characters is
typically of size $100\times32$, from which a feature sequence 25
frames can be generated. This length exceeds the lengths of most English
words. On top of that, the rectangular pooling windows yield rectangular
receptive fields (illustrated in Fig.~\ref{fig:recfield}), which
are beneficial for recognizing some characters that have narrow shapes,
such as 'i' and 'l'.

The network not only has deep convolutional layers, but also has recurrent
layers. Both are known to be hard to train. We find that the batch
normalization \cite{IoffeS15} technique is extremely useful for training
network of such depth. Two batch normalization layers are inserted
after the 5th and 6th convolutional layers respectively. With the
batch normalization layers, the training process is greatly accelerated.

We implement the network within the Torch7~\cite{Collobert11} framework,
with custom implementations for the LSTM units (in Torch7/CUDA), the
transcription layer (in C++) and the BK-tree data structure (in C++).
Experiments are carried out on a workstation with a 2.50 GHz Intel(R)
Xeon(R) E5-2609 CPU, 64GB RAM and an NVIDIA(R) Tesla(TM) K40 GPU.
Networks are trained with ADADELTA, setting the parameter $\rho$
to 0.9. During training, all images are scaled to $100\times32$ in
order to accelerate the training process. The training process takes
about 50 hours to reach convergence. Testing images are scaled to
have height 32. Widths are proportionally scaled with heights, but
at least 100 pixels. The average testing time is 0.16s/sample, as
measured on IC03 without a lexicon. The approximate lexicon search
is applied to the 50k lexicon of IC03, with the parameter $\delta$
set to 3. Testing each sample takes 0.53s on average.

\subsection{Comparative Evaluation\label{sec:evaluation}}

All the recognition accuracies on the above four public datasets,
obtained by the proposed CRNN model and the recent state-of-the-arts
techniques including the approaches based on deep models \cite{JaderbergVZ14,JaderbergSVZ14a,JaderbergSVZ14b},
are shown in Table~\ref{tbl:results}.

\begin{table*}[t]
\caption{Recognition accuracies (\%) on four datasets. In the second row, ``50'',
``1k'', ``50k'' and ``Full'' denote the lexicon used, and ``None''
denotes recognition without a lexicon. ({*}\cite{JaderbergSVZ14a}
is not lexicon-free in the strict sense, as its outputs are constrained
to a 90k dictionary.}

\vspace{0.2cm}

\footnotesize

\begin{centering}
\begin{tabular}{llccccccccccccc}
\hline 
\noalign{\vskip\doublerulesep}
 &  & \multicolumn{3}{c}{\textbf{IIIT5k}} &  & \multicolumn{2}{c}{\textbf{SVT}} &  & \multicolumn{4}{c}{\textbf{IC03}} &  & \textbf{IC13}\tabularnewline
\cline{3-5} \cline{7-8} \cline{10-13} \cline{15-15} 
\noalign{\vskip\doublerulesep}
 &  & \textbf{50} & \textbf{1k} & \textbf{None} &  & \textbf{50} & \textbf{None} &  & \textbf{50} & \textbf{Full} & \textbf{50k} & \textbf{None} &  & \textbf{None}\tabularnewline
\hline 
\noalign{\vskip\doublerulesep}
ABBYY \cite{WangBB11} &  & 24.3 & - & - &  & 35.0 & - &  & 56.0 & 55.0 & - & - &  & -\tabularnewline
\noalign{\vskip\doublerulesep}
Wang \emph{et al.} \cite{WangBB11} &  & - & - & - &  & 57.0 & - &  & 76.0 & 62.0 & - & - &  & -\tabularnewline
\noalign{\vskip\doublerulesep}
Mishra \emph{et al.} \cite{MishraAJ12} &  & 64.1 & 57.5 & - &  & 73.2 & - &  & 81.8 & 67.8 & - & - &  & -\tabularnewline
\noalign{\vskip\doublerulesep}
Wang \emph{et al.} \cite{WangWCN12} &  & - & - & - &  & 70.0 & - &  & 90.0 & 84.0 & - & - &  & -\tabularnewline
\noalign{\vskip\doublerulesep}
Goel \emph{et al.} \cite{GoelMAJ13} &  & - & - & - &  & 77.3 & - &  & 89.7 & - & - & - &  & -\tabularnewline
\noalign{\vskip\doublerulesep}
Bissacco \emph{et al.} \cite{BissaccoCNN13} &  & - & - & - &  & 90.4 & 78.0 &  & - & - & - & - &  & 87.6\tabularnewline
\noalign{\vskip\doublerulesep}
Alsharif and Pineau \cite{AlsharifP13} &  & - & - & - &  & 74.3 & - &  & 93.1 & 88.6 & 85.1 & - &  & -\tabularnewline
\noalign{\vskip\doublerulesep}
Almaz\'an \emph{et al.} \cite{AlmazanGFV14} &  & 91.2 & 82.1 & - &  & 89.2 & - &  & - & - & - & - &  & -\tabularnewline
\noalign{\vskip\doublerulesep}
Yao \emph{et al.} \cite{YaoBSL14} &  & 80.2 & 69.3 & - &  & 75.9 & - &  & 88.5 & 80.3 & - & - &  & -\tabularnewline
\noalign{\vskip\doublerulesep}
Rodr�guez-Serrano \emph{et al.} \cite{Rodriguez-Serrano15} &  & 76.1 & 57.4 & - &  & 70.0 & - &  & - & - & - & - &  & -\tabularnewline
\noalign{\vskip\doublerulesep}
Jaderberg \emph{et al.} \cite{JaderbergVZ14} &  & - & - & - &  & 86.1 & - &  & 96.2 & 91.5 & - & - &  & -\tabularnewline
\noalign{\vskip\doublerulesep}
Su and Lu \cite{SuL14} &  & - & - & - &  & 83.0 & - &  & 92.0 & 82.0 & - & - &  & -\tabularnewline
\noalign{\vskip\doublerulesep}
Gordo \cite{Gordo14} &  & 93.3 & 86.6 & - &  & 91.8 & - &  & - & - & - & - &  & -\tabularnewline
\noalign{\vskip\doublerulesep}
Jaderberg \emph{et al.} \cite{JaderbergSVZ14a} &  & 97.1 & 92.7 & - &  & 95.4 & 80.7{*} &  & \textbf{98.7} & \textbf{98.6} & 93.3 & \textbf{93.1}{*} &  & \textbf{90.8}{*}\tabularnewline
\noalign{\vskip\doublerulesep}
Jaderberg \emph{et al.} \cite{JaderbergSVZ14b} &  & 95.5 & 89.6 & - &  & 93.2 & 71.7 &  & 97.8 & 97.0 & 93.4 & 89.6 &  & 81.8\tabularnewline[\doublerulesep]
\hline 
\noalign{\vskip\doublerulesep}
CRNN &  & \textbf{97.6} & \textbf{94.4} & \textbf{78.2} &  & \textbf{96.4} & \textbf{80.8} &  & \textbf{98.7} & 97.6 & \textbf{95.5} & 89.4 &  & 86.7\tabularnewline
\hline 
\end{tabular}
\par\end{centering}

\label{tbl:results}
\end{table*}

In the constrained lexicon cases, our method consistently outperforms
most state-of-the-arts approaches, and in average beats the best text
reader proposed in \cite{JaderbergSVZ14a}. Specifically, we obtain
 superior performance on IIIT5k, and SVT compared to \cite{JaderbergSVZ14a},
only achieved lower performance on IC03 with the ``Full'' lexicon.
Note that the model in\cite{JaderbergSVZ14a} is trained on a specific
dictionary, namely that each word is associated to a class label.
Unlike \cite{JaderbergSVZ14a}, CRNN is not limited to recognize a
word in a known dictionary, and able to handle random strings (\emph{e.g.}
telephone numbers), sentences or other scripts like Chinese words.
Therefore, the results of CRNN are competitive on all the testing
datasets.

In the unconstrained lexicon cases, our method achieves the best performance
on SVT, yet, is still behind some approaches \cite{BissaccoCNN13,JaderbergSVZ14a}
on IC03 and IC13. Note that the blanks in the ``none'' columns of
Table~\ref{tbl:results} denote that such approaches are unable to
be applied to recognition without lexicon or did not report the recognition
accuracies in the unconstrained cases. Our method uses only synthetic
text with word level labels as the training data, very different to
PhotoOCR~\cite{BissaccoCNN13} which used 7.9 millions of real word
images with character-level annotations for training. The best performance
is reported by \cite{JaderbergSVZ14a} in the unconstrained lexicon
cases, benefiting from its large dictionary, however, it is not a
model strictly unconstrained to a lexicon as mentioned before. In
this sense, our results in the unconstrained lexicon case are still
promising.

\begin{table}[t]
\caption{Comparison among various methods. Attributes for comparison include:
1) being end-to-end trainable (E2E Train); 2) using convolutional
features that are directly learned from images rather than using hand-crafted
ones (Conv Ftrs); 3) requiring no ground truth bounding boxes for
characters during training (CharGT-Free); 4) not confined to a pre-defined
dictionary (Unconstrained); 5) the model size (if an end-to-end trainable
model is used), measured by the number of model parameters (Model
Size, M stands for millions).}

\footnotesize

\begin{centering}
\begin{tabular}{lccccc}
\noalign{\vskip\doublerulesep}
 & \begin{turn}{90}
E2E Train
\end{turn} & \begin{turn}{90}
Conv Ftrs
\end{turn} & \begin{turn}{90}
CharGT-Free
\end{turn} & \begin{turn}{90}
Unconstrained
\end{turn} & \begin{turn}{90}
Model Size
\end{turn}\tabularnewline
\hline 
\noalign{\vskip\doublerulesep}
Wang \emph{et al.} \cite{WangBB11} & \ding{55} & \ding{55} & \ding{55} & \ding{52} & -\tabularnewline
\noalign{\vskip\doublerulesep}
Mishra \emph{et al.} \cite{MishraAJ12} & \ding{55} & \ding{55} & \ding{55} & \ding{55} & -\tabularnewline
\noalign{\vskip\doublerulesep}
Wang \emph{et al.} \cite{WangWCN12} & \ding{55} & \ding{52} & \ding{55} & \ding{52} & -\tabularnewline
\noalign{\vskip\doublerulesep}
Goel \emph{et al.} \cite{GoelMAJ13} & \ding{55} & \ding{55} & \ding{52} & \ding{55} & -\tabularnewline
\noalign{\vskip\doublerulesep}
Bissacco \emph{et al.} \cite{BissaccoCNN13} & \ding{55} & \ding{55} & \ding{55} & \ding{52} & -\tabularnewline
\noalign{\vskip\doublerulesep}
Alsharif and Pineau \cite{AlsharifP13} & \ding{55} & \ding{52} & \ding{55} & \ding{52} & -\tabularnewline
\noalign{\vskip\doublerulesep}
Almaz\'an \emph{et al.} \cite{AlmazanGFV14} & \ding{55} & \ding{55} & \ding{52} & \ding{55} & -\tabularnewline
\noalign{\vskip\doublerulesep}
Yao \emph{et al.} \cite{YaoBSL14} & \ding{55} & \ding{55} & \ding{55} & \ding{52} & -\tabularnewline
\noalign{\vskip\doublerulesep}
Rodr�guez-Serrano \emph{et al.} \cite{Rodriguez-Serrano15} & \ding{55} & \ding{55} & \ding{52} & \ding{55} & -\tabularnewline
\noalign{\vskip\doublerulesep}
Jaderberg \emph{et al.} \cite{JaderbergVZ14} & \ding{55} & \ding{52} & \ding{55} & \ding{52} & -\tabularnewline
\noalign{\vskip\doublerulesep}
Su and Lu \cite{SuL14} & \ding{55} & \ding{55} & \ding{52} & \ding{52} & -\tabularnewline
\noalign{\vskip\doublerulesep}
Gordo \cite{Gordo14} & \ding{55} & \ding{55} & \ding{55} & \ding{55} & -\tabularnewline
\noalign{\vskip\doublerulesep}
Jaderberg \emph{et al.} \cite{JaderbergSVZ14a} & \ding{52} & \ding{52} & \ding{52} & \ding{55} & 490M\tabularnewline
\noalign{\vskip\doublerulesep}
Jaderberg \emph{et al.} \cite{JaderbergSVZ14b} & \ding{52} & \ding{52} & \ding{52} & \ding{52} & 304M\tabularnewline[\doublerulesep]
\hline 
\noalign{\vskip\doublerulesep}
CRNN & \ding{52} & \ding{52} & \ding{52} & \ding{52} & \textbf{8.3M}\tabularnewline
\hline 
\end{tabular}
\par\end{centering}

\label{tbl:methodcomp}
\end{table}

For further understanding the advantages of the proposed algorithm
over other text recognition approaches, we provide a comprehensive
comparison on several properties named E2E Train, Conv Ftrs, CharGT-Free,
Unconstrained, and Model Size, as summarized in Table~\ref{tbl:methodcomp}.

\textbf{E2E Train}: This column is to show whether a certain text
reading model is end-to-end trainable, without any preprocess or through
several separated steps, which indicates such approaches are elegant
and clean for training. As can be observed from Table~\ref{tbl:methodcomp},
only the models based on deep neural networks including~\cite{JaderbergSVZ14a,JaderbergSVZ14b}
as well as CRNN have this property.

\textbf{Conv Ftrs}: This column is to indicate whether an approach
uses the convolutional features learned from training images directly
or handcraft features as the basic representations.

\textbf{CharGT-Free}: This column is to indicate whether the character-level
annotations are essential for training the model. As the input and
output labels of CRNN can be a sequence, character-level annotations
are not necessary.

\textbf{Unconstrained}: This column is to indicate whether the trained
model is constrained to a specific dictionary, unable to handling
out-of-dictionary words or random sequences. Notice that though the
recent models learned by label embedding \cite{AlmazanGFV14,Gordo14}
and incremental learning \cite{JaderbergSVZ14a} achieved highly competitive
performance, they are constrained to a specific dictionary.

\textbf{Model Size}: This column is to report the storage space of
the learned model. In CRNN, all layers have weight-sharing connections,
and the fully-connected layers are not needed. Consequently, the number
of parameters of CRNN is much less than the models learned on the
variants of CNN \cite{JaderbergSVZ14a,JaderbergSVZ14b}, resulting
in a much smaller model compared with \cite{JaderbergSVZ14a,JaderbergSVZ14b}.
Our model has 8.3 million parameters, taking only 33MB RAM (using
4-bytes single-precision float for each parameter), thus it can be
easily ported to mobile devices.

Table~\ref{tbl:methodcomp} clearly shows the differences among different
approaches in details, and fully demonstrates the advantages of CRNN
over other competing methods.

\begin{figure}[h]
\begin{centering}
\includegraphics[width=0.7\linewidth]{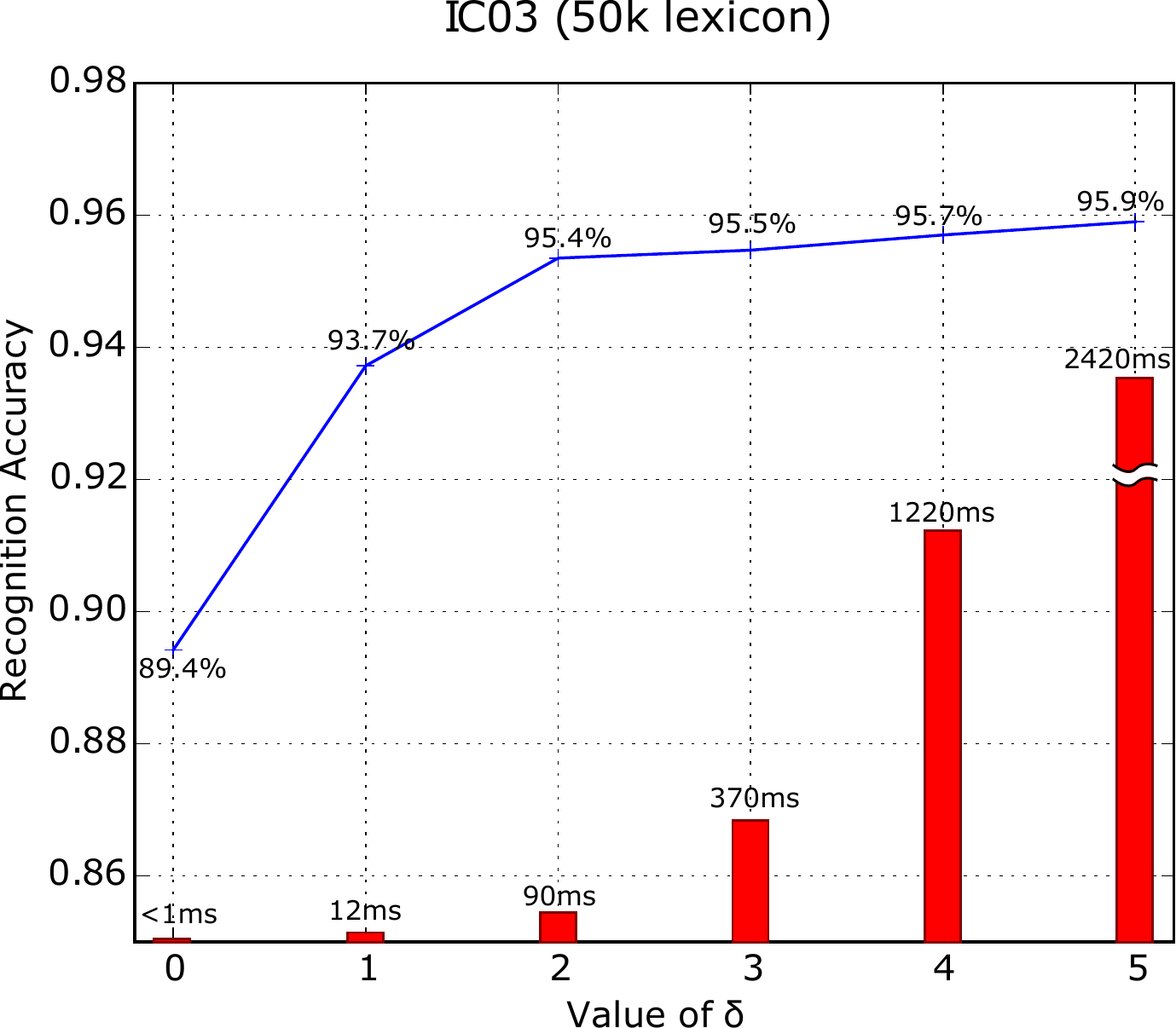}
\par\end{centering}

\caption{Blue line graph: recognition accuracy as a function parameter $\delta$.
Red bars: lexicon search time per sample. Tested on the IC03 dataset
with the 50k lexicon.}

\label{fig:impact-of-delta}
\end{figure}

In addition, to test the impact of parameter $\delta$, we experiment
different values of $\delta$ in Eq.~\ref{eq:largelex}. In Fig.~\ref{fig:impact-of-delta}
we plot the recognition accuracy as a function of $\delta$. Larger
$\delta$ results in more candidates, thus more accurate lexicon-based
transcription. On the other hand, the computational cost grows with
larger $\delta$, due to longer BK-tree search time, as well as larger
number of candidate sequences for testing. In practice, we choose
$\delta=3$ as a tradeoff between accuracy and speed.

\subsection{Musical Score Recognition\label{sec:musicalscore}}

A musical score typically consists of sequences of musical notes arranged
on staff lines. Recognizing musical scores in images is known as the
Optical Music Recognition (OMR) problem. Previous methods often requires
image preprocessing (mostly binirization), staff lines detection and
individual notes recognition~\cite{RebeloFPMGC12}. We cast the OMR
as a sequence recognition problem, and predict a sequence of musical
notes directly from the image with CRNN. For simplicity, we recognize
pitches only, ignore all chords and assume the same major scales (C
major) for all scores.

\begin{figure}[th]
\begin{centering}
\includegraphics[width=0.9\linewidth]{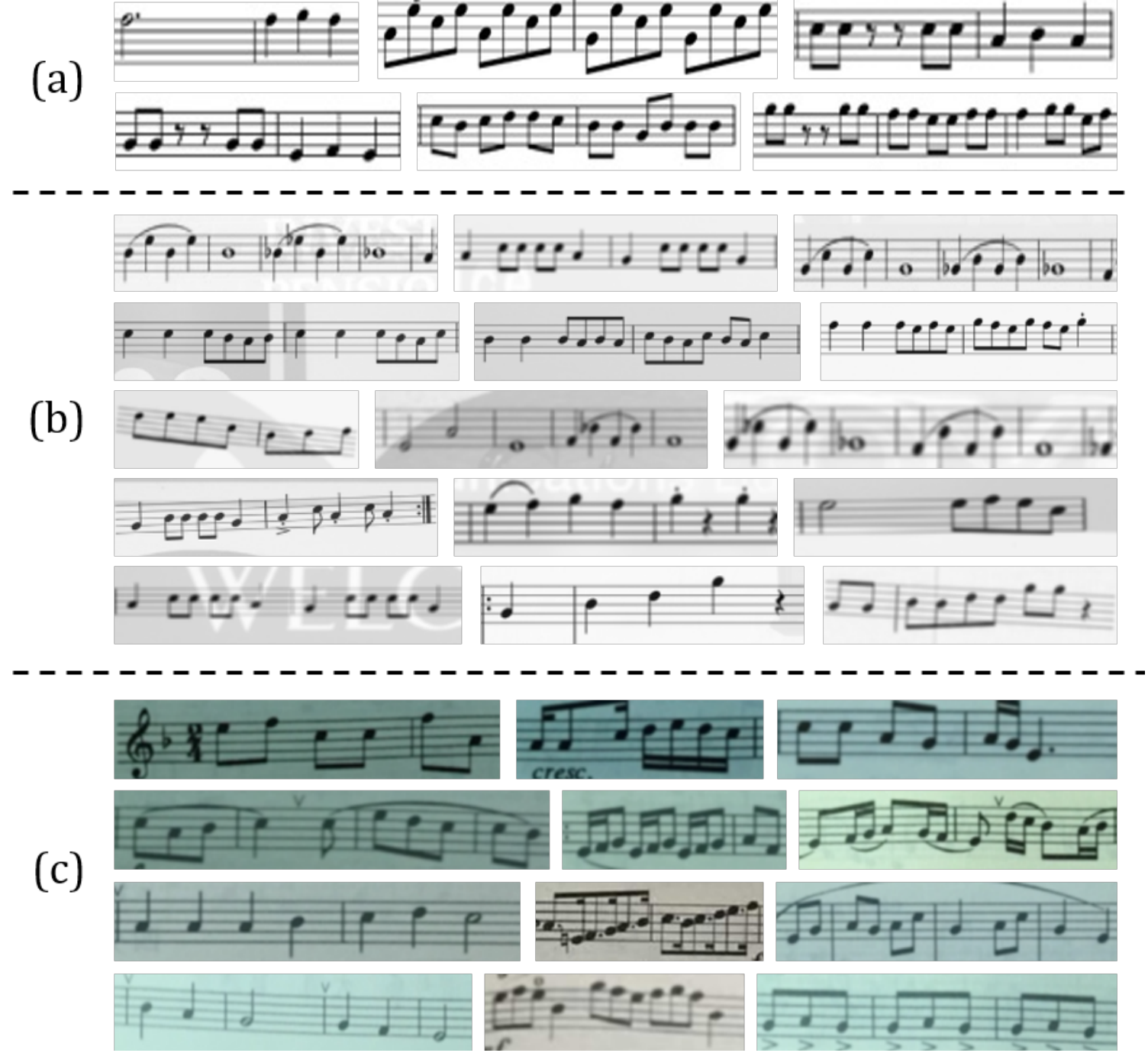}
\par\end{centering}

\caption{(a) Clean musical scores images collected from \cite{Musescore} (b)
Synthesized musical score images. (c) Real-world score images taken
with a mobile phone camera.}

\label{fig:musicsamples}
\end{figure}

To the best of our knowledge, there exists no public datasets for
evaluating algorithms on pitch recognition. To prepare the training
data needed by CRNN, we collect 2650 images from \cite{Musescore}.
Each image contains a fragment of score containing 3 to 20 notes.
We manually label the ground truth label sequences (sequences of not
ezpitches) for all the images. The collected images are augmented to
265k training samples by being rotated, scaled and corrupted with
noise, and by replacing their backgrounds with natural images. For
testing, we create three datasets: 1) ``Clean'', which contains
260 images collected from \cite{Musescore}. Examples are shown in
Fig.~\ref{fig:musicsamples}.a; 2) ``Synthesized'', which is created
from ``Clean'', using the augmentation strategy mentioned above.
It contains 200 samples, some of which are shown in Fig.~\ref{fig:musicsamples}.b;
3) ``Real-World'', which contains 200 images of score fragments
taken from music books with a phone camera. Examples are shown in
Fig.~\ref{fig:musicsamples}.c.\footnote{We will release the dataset for academic use.}

Since we have limited training data, we use a simplified CRNN configuration
in order to reduce model capacity. Different from the configuration
specified in Tab.~\ref{tbl:netconfig}, the 4th and 6th convolution
layers are removed, and the 2-layer bidirectional LSTM is replaced
by a 2-layer single directional LSTM. The network is trained on the
pairs of images and corresponding label sequences. Two measures are
used for evaluating the recognition performance: 1) fragment accuracy,
\emph{i.e.} the percentage of score fragments correctly recognized;
2) average edit distance, \emph{i.e.} the average edit distance between
predicted pitch sequences and the ground truths.  For comparison,
we evaluate two commercial OMR engines, namely the Capella~Scan~\cite{CapellaScan}
and the PhotoScore~\cite{PhotoScore}.

\begin{table}[h]
\caption{Comparison of pitch recognition accuracies, among CRNN and two commercial
OMR systems, on the three datasets we have collected. Performances
are evaluated by fragment accuracies and average edit distance (``fragment
accuracy/average edit distance'').}

\footnotesize

\begin{centering}
\begin{tabular}{lccc}
\noalign{\vskip\doublerulesep}
 & \textbf{Clean} & \textbf{Synthesized} & \textbf{Real-World}\tabularnewline
\hline 
\noalign{\vskip\doublerulesep}
Capella~Scan \cite{CapellaScan} & 51.9\%/1.75 & 20.0\%/2.31 & 43.5\%/3.05\tabularnewline
\noalign{\vskip\doublerulesep}
PhotoScore \cite{PhotoScore} & 55.0\%/2.34 & 28.0\%/1.85 & 20.4\%/3.00\tabularnewline
\noalign{\vskip\doublerulesep}
CRNN & \textbf{74.6\%/0.37} & \textbf{81.5\%/0.30} & \textbf{84.0\%/0.30}\tabularnewline
\hline 
\end{tabular}
\par\end{centering}

\label{tbl:musicaccuracy}
\end{table}

Tab.~\ref{tbl:musicaccuracy} summarizes the results. The CRNN outperforms
the two commercial systems by a large margin. The Capella~Scan and
PhotoScore systems perform reasonably well on the Clean dataset, but
their performances drop significantly on synthesized and real-world
data. The main reason is that they rely on robust binarization to
detect staff lines and notes, but the binarization step often fails
on synthesized and real-world data due to bad lighting condition,
noise corruption and cluttered background. The CRNN, on the other
hand, uses convolutional features that are highly robust to noises
and distortions. Besides, recurrent layers in CRNN can utilize contextual
information in the score. Each note is recognized not only itself,
but also by the nearby notes. Consequently, some notes can be recognized
by comparing them with the nearby notes, $e.g.$ contrasting their
vertical positions.

The results have shown the generality of CRNN, in that it can be readily
applied to other image-based sequence recognition problems, requiring
minimal domain knowledge. Compared with Capella~Scan and PhotoScore,
our CRNN-based system is still preliminary and misses many functionalities.
But it provides a new scheme for OMR, and has shown promising capabilities
in pitch recognition.

\section{Conclusion}

In this paper, we have presented a novel neural network architecture,
called Convolutional Recurrent Neural Network (CRNN), which integrates
the advantages of both Convolutional Neural Networks (CNN) and Recurrent
Neural Networks (RNN). CRNN is able to take input images of varying
dimensions and produces predictions with different lengths. It directly
runs on coarse level labels (\emph{e.g.} words), requiring no detailed
annotations for each individual element (\emph{e.g.} characters) in
the training phase. Moreover, as CRNN abandons fully connected layers
used in conventional neural networks, it results in a much more compact
and efficient model. All these properties make CRNN an excellent approach
for image-based sequence recognition.

The experiments on the scene text recognition benchmarks demonstrate
that CRNN achieves superior or highly competitive performance, compared
with conventional methods as well as other CNN and RNN based algorithms.
This confirms the advantages of the proposed algorithm. In addition,
CRNN significantly outperforms other competitors on a benchmark for
Optical Music Recognition (OMR), which verifies the generality of
CRNN.

Actually, CRNN is a general framework, thus it can be applied to other
domains and problems (such as Chinese character recognition), which
involve sequence prediction in images. To further speed up CRNN and
make it more practical in real-world applications is another direction
that is worthy of exploration in the future.

\section*{Acknowledgement}

This work was primarily supported by National Natural Science Foundation
of China (NSFC) (No. 61222308).

{\small
\bibliographystyle{ieee}
\bibliography{references}
}

\end{document}